\documentclass[times, review, 10pt]{elsarticle}
\usepackage{amssymb}
\usepackage{enumitem} % Add this to your preamble
\usepackage{algpseudocode}
\usepackage{algorithm}
\usepackage{pdflscape} % For landscape environment
\usepackage{algpseudocode}
\usepackage{amsmath}
\usepackage{lscape}
\usepackage{natbib}
\usepackage{multirow} 
\usepackage{amsmath}
\usepackage{rotating}
\usepackage{booktabs}
\usepackage{longtable}
\usepackage{array}
\usepackage{adjustbox}
\usepackage{geometry} % Adjust page margins if needed
\usepackage{adjustbox} % For scaling the table if necessary
\usepackage{mathrsfs}
\usepackage{hyperref}
\journal{journal}

\begin{document}

\begin{frontmatter}

%% Title, authors and addresses

%% use the tnoteref command within \title for footnotes;
%% use the tnotetext command for theassociated footnote;
%% use the fnref command within \author or \affiliation for footnotes;
%% use the fntext command for theassociated footnote;
%% use the corref command within \author for corresponding author footnotes;
%% use the cortext command for theassociated footnote;
%% use the ead command for the email address,
%% and the form \ead[url] for the home page:
%% \title{Title\tnoteref{label1}}
%% \tnotetext[label1]{}
%% \author{Name\corref{cor1}\fnref{label2}}
%% \ead{email address}
%% \ead[url]{home page}
%% \fntext[label2]{}
%% \cortext[cor1]{}
%% \affiliation{organization={},
%%            addressline={}, 
%%            city={},
%%            postcode={}, 
%%            state={},
%%            country={}}
%% \fntext[label3]{}

\title{Granular Ball K-Class Twin Support Vector Classifier} %% Article title

%% use optional labels to link authors explicitly to addresses:
%% \author[label1,label2]{}
%% \affiliation[label1]{organization={},
%%             addressline={},
%%             city={},
%%             postcode={},
%%             state={},
%%             country={}}
%%
%% \affiliation[label2]{organization={},
%%             addressline={},
%%             city={},
%%             postcode={},
%%             state={},
%%             country={}}
\cortext[cor1]{Corresponding author}
\author[inst2]{M. A. Ganaie \corref{cor1}}
\ead{mudasir@iitrpr.ac.in}
%\cortext[cor1]{Corresponding author}
% \affiliation[inst1]{organization={Department One},%Department and Organization
%             addressline={Address One}, 
%             city={City One},
%             postcode={00000}, 
%             state={State One},
%             country={Country One}}

\author[inst2]{Vrushank Ahire}
\ead{
2022csb1002@iitrpr.ac.in}
% \author[inst1,inst2]{Author Three}
\affiliation[inst2]{organization={Department of Computer Science and Engineering, Indian Institute of Technology Ropar},%Department and Organization
            % addressline={Address Two}, 
            city={Rupnagar},
            postcode={140001}, 
            state={Punjab},
            country={India}}
\author[inst3]{Anouck Girard}
\ead{
anouck@umich.edu}
% \author[inst1,inst2]{Author Three}
\affiliation[inst3]{organization={Department of Robotics, University of Michigan, Ann Arbor, USA},%Department and Organization
            % addressline={Address Two}, 
            city={Ann Arbor},
            postcode={48109-2140}, 
            state={Michigan},
            country={USA}}

%% Abstract
\begin{abstract}
%% Text of abstract
This paper introduces the Granular Ball K-Class Twin Support Vector Classifier (GB-TWKSVC), a novel multi-class classification framework that combines Twin Support Vector Machines (TWSVM) with granular ball computing. The proposed method addresses key challenges in multi-class classification by utilizing granular ball representation for improved noise robustness and TWSVM’s non-parallel hyperplane architecture solves two smaller quadratic programming problems, enhancing efficiency.
Our approach introduces a novel formulation that effectively handles multi-class scenarios, advancing traditional binary classification methods. Experimental evaluation on diverse benchmark datasets shows that GB-TWKSVC significantly outperforms current state-of-the-art classifiers in both accuracy and computational performance. The method's effectiveness is validated through comprehensive statistical tests and complexity analysis. Our work advances classification algorithms by providing a mathematically sound framework that addresses the scalability and robustness needs of modern machine learning applications. The results demonstrate GB-TWKSVC's broad applicability across domains including pattern recognition, fault diagnosis, and large-scale data analytics, establishing it as a valuable addition to the classification algorithm landscape.
\end{abstract}

%%Graphical abstract
% \begin{graphicalabstract}
% %\includegraphics{grabs}
% \end{graphicalabstract}

% %%Research highlights
% \begin{highlights}
% \item This work introduces a novel variant of Twin Support Vector Machines (TWSVM) that incorporates \textbf{granular ball computing} for \textbf{multi-class classification}, extending its application beyond traditional binary classification settings.
% \item By utilizing granular ball computing, the proposed method \textbf{clusters similar data points recursively} based on purity thresholds and label homogeneity, grouping them with other points of the same class to enhance \textbf{spatial locality}. This \textbf{reduces the likelihood of misclassification}, as points are more accurately classified within their own class, ultimately improving \textbf{overall classification efficiency}.
% \item The integration of granular ball computing with TWSVM significantly \textbf{accelerates the classification process by reducing computational time.} Experimental results show that the GB-TWKSVC model \textbf{outperforms state-of-the-art classifiers in terms of accuracy} across a wide range of benchmark datasets.
% \end{highlights}

%% Keywords
\begin{keyword}
%% keywords here, in the form: keyword \sep keyword

%% PACS codes here, in the form: \PACS code \sep code

%% MSC codes here, in the form: \MSC code \sep code
%% or \MSC[2008] code \sep code (2000 is the default)

Multi-class Classification, Twin Support Vector Machine, Granular Ball Computing, Noise Resistance

\end{keyword}

\end{frontmatter}

%% Add \usepackage{lineno} before \begin{document} and uncomment 
%% following line to enable line numbers
%% \linenumbers

%% main text
%%

%% Use \section commands to start a section

\section{Introduction}

The field of machine learning witnessed remarkable advancements in classification algorithms over the past few decades, with Support Vector Machines (SVMs) emerging as powerful tools for solving complex classification problems. Since their introduction by Vapnik \cite{vapnik2013nature}, SVMs gain popularity due to their ability to handle high-dimensional data and their strong theoretical foundations in statistical learning theory. However, the original SVM formulation is limited to binary classification, prompting researchers to explore methods for handling multi-class problems.

The journey towards multi-class SVMs begins with strategies like the ``one-versus-one" approach proposed by Hastie and Tibshirani \cite{hastie1997classification}. This method involves training binary classifiers for each pair of classes and combining their decisions. Building on this concept, Angulo and Català introduce K-SVCR \cite{angulo2000k}, a multi-class SVM that assigns outputs of +1, -1, or 0 to training patterns, enhancing the model's ability to handle complex class relationships.

A significant breakthrough comes with the introduction of Twin Support Vector Machines (TWSVM) by Khemchandani et al. \cite{khemchandani2007twin}. TWSVM aims to improve computational efficiency by solving two smaller quadratic programming problems instead of a single large one. This innovation is particularly effective for large-scale datasets and sparks a new line of research focused on improving and extending the TWSVM framework.

The concept of multi-class TWSVM begins to take shape, with researchers exploring various strategies to extend the binary TWSVM to handle multiple classes simultaneously. These approaches offer different trade-offs between computational complexity and classification accuracy. For example, Xu et al. propose the Twin Multi-Class Classification Support Vector Machine \cite{xu2013twin}, which extends the binary TWSVM to handle multiple classes efficiently.

As research in this area intensifies, several key themes emerge: 
\begin{itemize} 
    \item Computational Efficiency: A major focus is on improving the speed and scalability of multi-class SVMs, particularly for large-scale datasets. The Least Squares Twin Multi-Class Classification Support Vector Machine (LSTKSVC) introduced by Nasiri et al. \cite{nasiri2015least} replaces quadratic programming problems with systems of linear equations, significantly reducing computational complexity. This approach is later refined by Ali et al. \cite{ali2022regularized} with the addition of regularization to address overfitting issues.
    \item Incorporation of Local Information: Researchers recognize the importance of utilizing local data structures to enhance classification accuracy. Xu's K-nearest neighbor-based weighted multi-class twin support vector machine \cite{xu2016k} uses weight matrices to exploit local information within classes. This idea is further developed by Tanveer et al. \cite{tanveer2021least}, combining efficiency with improved handling of imbalanced datasets. 
    \item Robustness and Outlier Handling: As multi-class SVMs are applied to more complex real-world problems, the need for robust methods that can handle noisy data and outliers becomes apparent. Qiang et al.'s TSVM-M3 \cite{qiang2022tsvm} incorporates multi-order moment matching to reduce sensitivity to outliers in large-scale multi-class classification. Their robust weighted linear loss twin multi-class support vector regression method \cite{qiang2020robust} further addresses these challenges.

    \item Integration with Deep Learning: Recent years see efforts to combine the strengths of SVMs with deep learning techniques. Xie et al.'s deep multi-view multiclass twin support vector machines \cite{xie2023deep} demonstrate the potential of integrating TWSVM with deep neural networks for improved performance on multi-view data.
\end{itemize}

Parallel to these developments, the concept of granular computing begins to gain traction in the machine-learning community. The Granular-Ball Support Vector Machine (GBSVM) introduced by Xia et al. \cite{xia2022gbsvm} represents a significant departure from traditional point-based SVMs by using granular balls as input. This approach offers improved robustness and efficiency, particularly for noisy and large datasets.

Our proposed Granular Ball K-Class Twin Support Vector Machine (GB-TWKSVC) builds upon these foundations, combining the strengths of multi-class TWSVM with the robust data representation offered by granular ball computing. It addresses several key challenges in multi-class classification:
\begin{itemize}
    \item Computational Efficiency: By using the TWSVM framework and granular ball representation, GB-TWKSVC is designed to handle large-scale datasets efficiently.
   \item Robustness: The use of granular balls, formed through hierarchical k-means clustering, enhances the model's ability to handle complex data, improving resilience to outliers and data perturbations, with minimal reliance on explicit experimentation with noisy data.
    \item Multi-class Handling: GB-TWKSVC extends the binary classification capabilities of TWSVM to efficiently handle multi-class problems without resorting to multiple binary classifiers.
    \item Improved Generalization: By incorporating ideas from recent advancements in the field, such as regularization and local information exploitation, GB-TWKSVC aims to achieve better generalization performance across a wide range of datasets.
\end{itemize}

The development of GB-TWKSVC is also informed by advancements in related areas, such as the application of TSVM to regression problems \cite{peng2010tsvr} and the use of transductive bounds for multi-class majority vote classifiers \cite{feofanov2019transductive}. These related works provide valuable insights into the theoretical foundations and practical considerations of multi-class SVM variants.

Through extensive experiments, we demonstrate that GB-TWKSVC offers superior performance in terms of accuracy overall. Even after the creation of \( O(K^2) \) classifiers, similar to the Twin-KSVC with the ``1-versus-1-versus-rest" approach, GB-TWKSVC outperforms Twin-KSVC. Additionally, despite 1-versus-rest OTSVM employing \( O(K) \) classifiers, GB-TWKSVC remains competitive in terms of computational efficiency. Our approach builds upon the comprehensive review of multi-class TWSVM methods conducted by Ding et al. \cite{ding2019review}, addressing identified areas for future research and incorporating recent advancements in the field.

The rest of this paper is organized as follows: Section 2 provides a detailed background on TWSVM and granular ball computing, explaining the fundamental concepts and their relevance to our proposed method. Section 3 introduces the GB-TWKSVC algorithm, detailing its formulation and theoretical foundations. Section 4 presents our experimental results, offering a comprehensive comparison with state-of-the-art methods across various datasets and performance metrics. Finally, Section 5 concludes the paper and discusses future research directions, including potential extensions to incorporate ideas from this paper such as the development of online and incremental learning variants for streaming data applications.

Through GB-TWKSVC, we aim to contribute to the ongoing evolution of multi-class classification algorithms, offering a robust and efficient solution that addresses the challenges of modern machine-learning tasks. By combining the strengths of TWSVM, granular ball computing, and recent advancements in multi-class classification, we believe our method represents a significant step forward in the field of machine learning and pattern recognition.

\section{Related Work}
\label{sec:Preliminaries}
This section examines Twin Support Vector Machine (TSVM) models that utilize granular ball data inputs. Additionally, we discuss multi-class methods founded on two distinct approaches: the ``one-versus-rest" paradigm, which employs K binary Support Vector Machine (SVM) classifiers, and the ``one-versus-one-versus-rest" structure. The latter approach forms the basis for both K-SVCR (Support Vector Classification Regression for K-class classification) and Twin-KSVC models.

\subsection{Twin Support Vector Machines (TSVM)}

Twin Support Vector Machines (TSVM) represent an advancement in binary classification algorithms, offering a distinct approach from traditional Support Vector Machines (SVM). While conventional SVMs seek a single optimal hyperplane to separate classes, TSVMs employ a pair of non-parallel hyperplanes, each aligned closely with one class while maintaining separation from the other.
The TSVM framework offers several key advantages, including computational efficiency and improved classification accuracy, particularly for non-linearly separable or imbalanced datasets. By decomposing the classification problem into two smaller quadratic programming problems (QPPs), TSVMs achieve faster training times compared to standard SVMs, especially for large-scale datasets, approximately 4 times faster with a complexity of 2 × O((n/2)³).

Mathematically, given a binary classification problem with a training dataset $\mathcal{D} = {(\mathbf{z}_i, y_i) \mid i = 1, 2, \ldots, N}$, where $\mathbf{z}_i \in \mathbb{R}^d$ represents the $i$-th feature vector and $y_i \in {-1, +1}$ denotes its corresponding class label, TSVM aims to find two non-parallel hyperplanes:

\begin{equation}
z^T {\omega}^{(1)} + b^{(1)} = 0 \quad \text{and} \quad z^T {\omega}^{(2)} + b^{(2)} = 0
\end{equation}
where ${\omega}^{(1)}, {\omega}^{(2)} \in \mathbb{R}^d$ are weight vectors and $b^{(1)}, b^{(2)} \in \mathbb{R}$ are bias terms.
Let ${A} \in \mathbb{R}^{n_1 \times d}$ and ${B} \in \mathbb{R}^{n_2 \times d}$ be matrices whose rows are the feature vectors of the positive and negative classes, respectively, where $n_1 + n_2 = N$. 

The primal optimization problems for TSVM can be formulated as:

\begin{equation}
\begin{aligned}
\min_{\omega^{(1)}, b^{(1)}, \xi} \quad & \frac{1}{2} \|A \omega^{(1)} + e_1 b^{(1)}\|^2 + c_1 e_2^T \xi, \\
\text{s.t.} \quad & - (B \omega^{(1)} + e_2 b^{(1)}) + \xi \geq e_2, &\xi \geq 0,
\end{aligned}
\\
\end{equation}
\begin{center}
and
\end{center}
\begin{equation}
\begin{aligned}
\min_{\omega^{(2)}, b^{(2)}, \eta} \quad & \frac{1}{2} \|B \omega^{(2)} + e_2 b^{(2)}\|^2 + c_2 e_1^T \eta, \\
\text{s.t.} \quad & - (A \omega^{(2)} + e_1 b^{(2)}) + \eta \geq e_1, &\eta \geq 0,
\end{aligned}
\\
\end{equation}
Here, $\mathbf{e}_1 \in \mathbb{R}^{n_1}$ and $\mathbf{e}_2 \in \mathbb{R}^{n_2}$ are vectors of ones, $c_1$ and $c_2$ are regularization parameters, and ${\xi}$ and ${\eta}$ are slack variables.

The dual formulation of these problems leads to more efficient solutions:

\begin{equation}
\begin{aligned}
\max_{\alpha} \quad & e_2^T \alpha - \frac{1}{2} \alpha^T G (H^T H)^{-1} G^T \alpha, \\
\text{s.t.} \quad & 0 \leq \alpha \leq c_1 e_2,
\end{aligned}
\label{eq:dual_problem_1}
\end{equation}

\begin{center}
    and
\end{center}
\begin{equation}
\begin{aligned}
\max_{\beta} \quad & e_1^T \beta - \frac{1}{2} \beta^T P (Q^T Q)^{-1} P^T \beta, \\
\text{s.t.} \quad & 0 \leq \beta \leq c_2 e_1.
\end{aligned}
\label{eq:dual_problem_2}
\end{equation}
where  $H = [A \; e_1]$, $G = [B \; e_2]$, $P = [A \; e_1]$, and $Q = [B \; e_2]$.\\
The solutions $\alpha$ and $\beta$ from the dual problems provide the parameters for the hyperplanes as:

\begin{equation}
\begin{aligned}
v &= -(H^T H)^{-1} G^T \alpha, \quad \text{where} \quad v = \begin{bmatrix} \omega^{(1)} \\ b^{(1)} \end{bmatrix}, \\[6pt]
v^* &= -(Q^T Q)^{-1} P^T \beta, \quad \text{where} \quad v^* = \begin{bmatrix} \omega^{(2)} \\ b^{(2)} \end{bmatrix}.
\end{aligned}
\end{equation}
For a new data point $\mathbf{z}_{\text{new}}$, the classification decision is made based on the proximity to the two hyperplanes:
\begin{equation}
    \text{class}(\mathbf{z}_{\text{new}}) = \text{sign}\left(\frac{|z_{new}^T {\omega}^{(1)} + b^{(1)}|}{\|\mathbf{\omega}^{(1)}\|} - \frac{|z_{new}^T {\omega}^{(2)} + b^{(2)}|}{\|\mathbf{\omega}^{(2)}\|}\right)
\end{equation}
where sign(·) represents the signum function. This rule assigns the new point to the class corresponding to the nearest hyperplane.

\subsection{Granular Ball Support Vector Machines (GB-SVM)}
Granular Ball Support Vector Machines (GB-SVM), introduced by Xia et al. \cite{xia2022gbsvm}, enhance the traditional SVM framework by representing data points as granular balls instead of discrete points. Each granular ball is defined by a center \( c_i \) and radius \( r_i \), where \( c_i \) represents the mean of enclosed data points and \( r_i \) captures their variability. This representation enables robust classification, particularly in scenarios with noisy data or outliers.

The total number of data points in the dataset is \( n \), but there are \( m \) granular balls, each represented by a center \( c_i \) and a radius \( r_i \), where \( i = 1, \dots, m \).

The fundamental constraint for support planes \( l'_1 \) and \( l'_2 \) incorporating granular balls is:
\begin{equation}
    y_i \omega \cdot c_i + y_i b - \|\omega\| r_i \geq 1
    \label{eq:constraint}
\end{equation}

The support planes are defined as:
\begin{equation}
\begin{aligned}
l'_1 &: \omega \cdot c_i - \|\omega\| r_i + b = 1, \quad y_i = +1 \\
l'_2 &: \omega \cdot c_i + \|\omega\| r_i + b = -1, \quad y_i = -1 \\
l_0' &: \omega \cdot c_i + b = 0
\end{aligned}
\end{equation}

The goal is to find the best separation hyperplane \( (\omega, b) \) that takes the granular balls into consideration. 

For inseparable cases, the primal optimization problem with slack variables \( \xi_i \) and penalty coefficient \( C \) is formulated as:
\begin{equation}
\begin{aligned}
& \underset{\omega, b, \xi}{\text{min}}
& & \frac{1}{2} \|\omega\|^2 + C \sum_{i=1}^{m} \xi_i \\
& \text{subject to}
& & y_i (\omega \cdot c_i + b) - \|\omega\| r_i \geq 1 - \xi_i, \\
& & & \xi_i \geq 0, \quad i = 1, \dots, m,
\end{aligned}
\vspace{0.3cm}
\end{equation}

The corresponding dual formulation introduces Lagrange multipliers \( \alpha_i \):
\begin{equation}
\begin{aligned}
& \underset{\alpha}{\text{max}}
& & -\frac{1}{2} P^2 - \frac{1}{2} Q^2 + \|P\| Q + \sum_{i=1}^{m} \alpha_i \\
& \text{subject to}
& & \quad\sum_{i=1}^{m} \alpha_i y_i = 0, \\
& & & 0 \leq \alpha_i \leq C, \quad i = 1, \dots, m,
\end{aligned}
\vspace{0.3cm}
\end{equation}

where \( P = \sum_{i=1}^{m} \alpha_i y_i c_i \) and \( Q = \sum_{i=1}^{m} \alpha_i r_i \). \( P \) is a vector and \( \|P\| \) is its norm (scalar). The optimal weight vector \( \omega \) is obtained as:

\begin{equation}
\omega = \frac{(\|P\| - Q)P}{\|P\|}
\end{equation}
given:
\begin{equation}
\|\omega\| = \|P\| - Q
\end{equation}

The bias term (\( b \))  is computed by averaging over all support vectors:

\begin{equation}
b = \frac{1}{n_s} \sum_{i=1}^{n_s} \left( y_i - \omega \cdot c_i + y_i \|\omega\| r_i \right)
\end{equation}

where \( n_s \) denotes the number of support vectors in the SVM model.

This granular ball representation enhances the model's robustness to noise and outliers, providing a foundation for advanced classification approaches that incorporate additional contextual information.

\subsection{K-SVCR}

The problem of multi-class classification remains a significant challenge in the domain of machine learning, particularly when utilizing support vector machines (SVMs), which are inherently designed for binary classification. A widely adopted strategy to address this challenge involves decomposing the multi-class problem into multiple binary classification tasks, followed by a reconstruction phase to synthesize the outcomes. Angulo et al. \cite{angulo2000k} proposed the Support Vector Classification-Regression (K-SVCR) algorithm, which provides an innovative approach to multi-class classification through a distinctive decomposition-reconstruction framework.

The K-SVCR algorithm employs a ``1-versus-1-versus-rest" scheme during the decomposition phase. Unlike conventional approaches, this method integrates both classification and regression elements within the SVM framework. The algorithm is designed to handle $K$ classes by training machines with ternary outputs $\{-1, 0, +1\}$, where $-1$ and $+1$ represent the two classes being explicitly separated, and $0$ represents all other classes. For a training set $\mathcal{T} = \{(z_p, y_p)\}_{p=1}^n \subset \mathcal{Z} \times \mathcal{Y}$, where $\mathcal{Y} = \{\omega_1, \dots, \omega_K\}$ and $n$ represents the total number of training samples, K-SVCR aims to find a decision function $f(z, \omega)$ that maps inputs to one of three outputs:

\begin{equation}
    f(z_p) = \begin{cases}
    +1, & p = 1, \dots, n_1 \\
    -1, & p = n_1 + 1, \dots, n_1 + n_2 \\
    0, & p = n_1 + n_2 + 1, \dots, n
    \end{cases}
\end{equation}
The K-SVCR algorithm extends the standard SVM binary classification framework by incorporating elements from SVM regression. For binary classification, the optimization problem is typically formulated as follows:
\begin{equation}
\begin{aligned}
\text{arg} \min_{\omega, b, \xi} \quad \frac{1}{2} \|\omega\|^2_\mathcal{F} + c_1 \sum_{i=1}^{n} \xi_i \\
\text{s.t.} \quad y_i \cdot (\langle \omega, \Phi(z_i) \rangle + b) &\geq 1 - \xi_i, \quad i = 1, \dots, n
\end{aligned}
\end{equation}
where $\xi_i \geq 0$ are slack variables. Here, $\mathcal{F}$ represents the feature space where $\omega \in \mathcal{F}$, $b \in \mathbb{R}$, and $\Phi: \mathcal{Z} \to \mathcal{F}$ denotes the mapping of input data into the feature space.

% K-SVCR generalizes this formulation by incorporating the $\epsilon$-insensitive loss function commonly used in SVM regression. The resulting optimization problem for K-SVCR is expressed as:
% \begin{equation}
%     \text{arg} \min_{\omega, b, \xi, \eta, \eta^*} \, \frac{1}{2} \|\omega\|^2_F + C \sum_{i=1}^{l} \xi_i + D \sum_{i=1}^{l} (\eta_i + \eta^*) \\
% \end{equation}

% \begin{equation}
% \begin{aligned}
% \text{s.t.} \quad \quad y_i \cdot (\langle \omega, \Phi(z_i) \rangle + b) &\geq 1 - \xi_i, \quad i = 1, \dots, n_{12} \\
% -\epsilon - \eta^*_i \leq \langle \omega, \Phi(z_i) \rangle + b &\leq \epsilon + \eta_i, \quad i = n_{12} + 1, \dots, n
% \end{aligned}
% \end{equation}
% where \(\xi_i\), \(\eta_i\), \(\eta^*_i \geq 0\) are slack variables, and \(\epsilon\) is a pre-chosen positive parameter constrained to be less than 1 to prevent class overlap.

% The dual formulation of this problem leads to:
% \begin{equation}
%     \arg\min L(\alpha) = \frac{1}{2} \alpha^T \cdot H \cdot \alpha + c^T \cdot \alpha    
% \end{equation}
% where $H$ is the kernel matrix and $c$ is a vector of constants.

% This formulation allows K-SVCR to simultaneously handle binary classification and regression-like treatment for the remaining classes. The reconstruction phase employs a specialized voting scheme considering both positive and negative votes, enhancing the algorithm's fault tolerance.

K-SVCR generalizes this formulation by incorporating the $\epsilon$-insensitive loss function commonly used in SVM regression. The resulting optimization problem for K-SVCR is expressed as:

\begin{equation}
\begin{aligned}
    \text{arg} \min_{\omega, b, \xi, \eta, \eta^*} \, & \frac{1}{2} \|\omega\|^2_\mathcal{F} + C \sum_{i=1}^{n} \xi_i + D \sum_{i=1}^{n} (\eta_i + \eta^*) \\
    \text{s.t.} \quad & y_i \cdot (\langle \omega, \Phi(z_i) \rangle + b) \geq 1 - \xi_i, \quad i = 1, \dots, n_{12} \\
    & -\epsilon - \eta^*_i \leq \langle \omega, \Phi(z_i) \rangle + b \leq \epsilon + \eta_i, \quad i = n_{12} + 1, \dots, n
\end{aligned}
\end{equation}

where $\xi_i$, $\eta_i$, $\eta^*_i \geq 0$ are slack variables, and $\epsilon$ is a pre-chosen positive parameter constrained to be less than 1 to prevent class overlap. The parameters $C$ and $D$ are regularization constants. $C$ controls the trade-off between maximizing the margin and minimizing the classification error, while $D$ balances the regression-like treatment of the remaining classes.

The dual formulation of this problem can be expressed as:

% \begin{equation}
%     \text{arg} \min L(\alpha) = \frac{1}{2} \alpha^T H \alpha + c^T \alpha
% \end{equation}
% where $\alpha = (\alpha_1, \dots, \alpha_{n_{12}}, \alpha_{n_{12}+1}, \dots, \alpha_{n}) \in \mathbb{R}^{n_{12} + n_3 + n_3}$ and $c^T = \left( -\frac{1}{y_1}, \dots, -\frac{1}{y_{n_{12}}}, \dots \right) \in\mathbb{R}^{n_{12} + n_3 + n_3}$. The kernel matrix $H = (k(x_i, x_j))$ is defined by the dot products in the transformed feature space $\mathcal{F}$ and satisfies $H = H^T \in M(\mathbb{R}^{n_{12} + n_3 + n_3}; \mathbb{R}^{n_{12} + n_3 + n_3})$.

\begin{equation}
    \text{arg} \min L(\alpha) = \frac{1}{2} \alpha^T H \alpha + c^T \alpha
\end{equation}
where
\begin{align}
    \alpha &= (\alpha_1, \dots, \alpha_{n_{12}}, \alpha_{n_{12}+1}, \dots, \alpha_{n}) \in \mathbb{R}^{n_{12} + n_3 + n_3}, \\
    c^T &= \left( -\frac{1}{y_1}, \dots, -\frac{1}{y_{n_{12}}}, \dots \right) \in \mathbb{R}^{n_{12} + n_3 + n_3}.
\end{align}
The kernel matrix \( H = (k(x_i, x_j)) \) is defined by the dot products in the transformed feature space \( \mathcal{F} \) and satisfies
\begin{equation}
    H = H^T \in M\left(\mathbb{R}^{n_{12} + n_3 + n_3}; \mathbb{R}^{n_{12} + n_3 + n_3}\right).
\end{equation}

This formulation allows K-SVCR to simultaneously handle binary classification and regression-like treatment for the remaining classes. The reconstruction phase employs a voting scheme considering both positive and negative votes, enhancing the algorithm's fault tolerance.

\subsection{Twin-KSVC}

Twin k-class Support Vector Machines (Twin-KSVC) \cite{xu2013twin} was proposed as an extension to traditional support vector machines for multi-class classification tasks. This approach employs a ``1-versus-1-versus-rest" structure, wherein two distinct sample sets are selected from k classes and treated as focal partitions. The algorithm maps the remaining samples into an intermediate region between these non-parallel hyperplanes, resulting in a ternary output system \{-1, 0, +1\}. A key advantage of Twin-KSVC over K-SVCR is its ability to resolve a pair of smaller-sized quadratic programming problems (QPPs) instead of a single large one, potentially improving computational efficiency.

Let $A \in \mathbb{R}^{n_1 \times d}$ and $B \in \mathbb{R}^{n_2 \times d}$ denote the two focal sample sets, labeled +1 and -1 respectively. The remaining samples are represented by $C \in \mathbb{R}^{n_3 \times d}$ and labeled 0. The objective is to generate two non-parallel hyperplanes:

\begin{equation}
z^T w^{(1)} + b^{(1)} = 0 \quad \text{and} \quad z^T w^{(2)} + b^{(2)} = 0
\end{equation}
These hyperplanes are obtained by solving the following pair of QPPs:
\begin{equation}
\begin{aligned}
\min_{w^{(1)}, b^{(1)}, \xi, \eta} & \frac{1}{2} \|Aw^{(1)} + e_1b^{(1)}\|^2 + c_1e_2^T\xi + c_2e_3^T\eta \\
\text{s.t.} \quad & -(Bw^{(1)} + e_2b^{(1)}) + \xi \geq e_2, & \xi \geq 0,\\
& -(Cw^{(1)} + e_3b^{(1)}) + \eta \geq (1-\epsilon)e_3, & \eta \geq 0 \\
\end{aligned}
\end{equation}
\begin{center}
and
\end{center}
\begin{equation}
\begin{aligned}
\min_{w^{(2)}, b^{(2)}, \xi^*, \eta^*} & \frac{1}{2} \|Bw^{(2)} + e_2b^{(2)}\|^2 + c_3e_1^T\xi^* + c_4e_3^T\eta^* \\
\text{s.t.} \quad & (Aw^{(2)} + e_1b^{(2)}) + \xi^* \geq e_1, & \xi^* \geq 0, \\
& (Cw^{(2)} + e_3b^{(2)}) + \eta^* \geq (1-\epsilon)e_3, 
& \eta^* \geq 0
\end{aligned}
\end{equation}
where $w^{(1)}, w^{(2)} \in \mathbb{R}^{d \times 1}$, $b^{(1)}, b^{(2)} \in \mathbb{R}$, $\xi \in \mathbb{R}^{n_2 \times 1}$, $\eta \in \mathbb{R}^{n_3 \times 1}$, $\xi^* \in \mathbb{R}^{n_1 \times 1}$, $\eta^* \in \mathbb{R}^{n_3 \times 1}$, and $e_i$ denotes a vector of ones with appropriate dimensions.

The dual problem of the first QPP can be formulated as:
\begin{equation}
\begin{aligned}
\max_c &\quad -\frac{1}{2}c^T N(H^T H)^{-1} N^T c + e_4^T c \\
\text{s.t.}  & \quad 0\leq c \leq F
\end{aligned}
\end{equation}
where $H = [A \, e_1]$, $G = [B \, e_2]$, $M = [C \, e_3]$, $N = [G;M]$, $c = [\alpha;\beta]$, $e_4 = [e_2;e_3(1-\epsilon)]$, and $F = [c_1e_2; c_2e_3]$.\\
The solution vector $u = [w^{(1)}; b^{(1)}]$ can be obtained as:
\begin{equation}
u = (H^T H + \delta I_{d+1})^{-1} (G^T \alpha + M^T \beta)
\end{equation}
where $\delta$ is a small positive regularization term introduced in the case when the matrix is (nearly) singular.

Similarly, we can derive the dual problem of QPP, and the other solution vector $u^*$ can be obtained as:
\begin{equation}
u^* = (G^T G + \delta I_{d+1})^{-1} (H^T \alpha^* + M^T \beta^*)
\end{equation}
This formulation of Twin-KSVC provides a great approach to multi-class classification, offering potential advantages in terms of computational efficiency and classification accuracy for complex datasets.

\section{Proposed Model: GB-TWKSVC }

The performance of traditional classification methods, including classical Twin K-class Support Vector Machines, can be limited by challenges such as sensitivity to label noise and the inability to handle complex, multi-dimensional data effectively. While these models are powerful in binary classification tasks, they struggle to generalize in multi-class scenarios, which are prevalent in real-world applications. Moreover, these methods often face scalability issues that hinder their adaptability to various types of data distributions. Therefore, there is a need for more robust, scalable, and noise-resilient classification techniques.

Granular computing provides a promising solution, especially through the concept of granular balls. While its success has been demonstrated in binary classification, its potential in multi-class problems has yet to be fully explored. By grouping similar data points into granular balls, the proposed approach introduces an additional layer of abstraction that improves generalization. This is in line with human cognitive processes, where information is often abstracted into meaningful chunks or granules. This abstraction enhances the interpretability of the models and helps reduce computational complexity by working with clusters of data rather than individual points.

The granular ball representation also facilitates a transformation of the feature space, potentially uncovering hidden patterns that are not obvious in the original data. The adaptability of granular balls to varying data densities and distributions allows for more flexible decision boundaries compared to traditional hyperplane-based methods. This flexibility is especially useful in addressing class imbalance, a common challenge in many classification tasks. By integrating granular ball computing with Twin Support Vector Machines, our GB-KTVSM model overcomes the limitations of existing methods, offering a more efficient and adaptable multi-class classification framework that enhances performance across a variety of real-world applications.

\subsection{Model Overview}
This section provides an overview of the model's key components and processes.
\begin{figure}[h]
\centering
\includegraphics[width=0.8\textwidth]{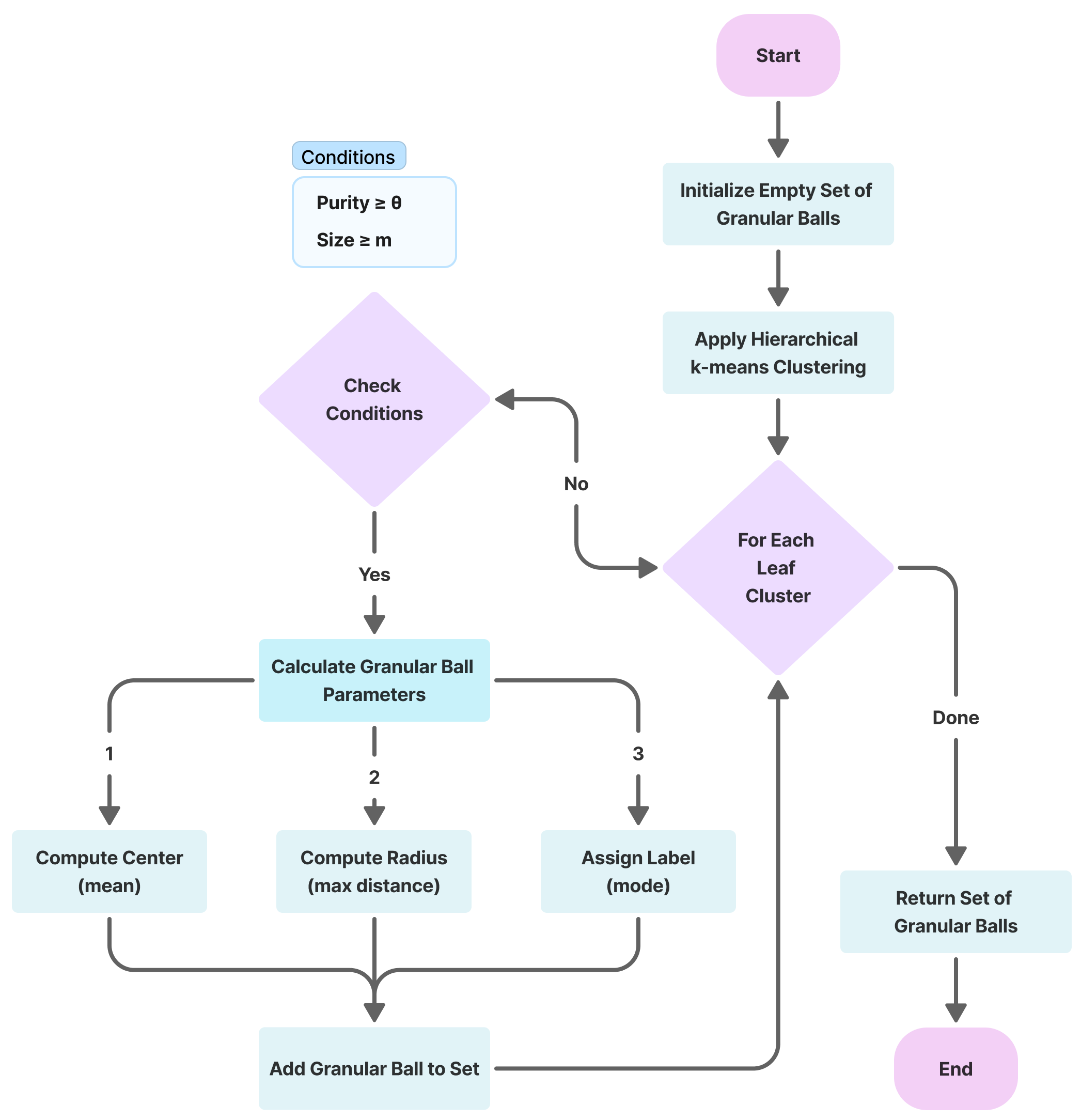}
\caption{Granular Ball Generation Flowchart}
\label{fig:generation_flow}
\end{figure}
\subsubsection{Granular Ball Generation}
The initial step in the proposed model involves the generation of granular balls using a hierarchical k-means clustering algorithm. The granular ball generation process can be formalized, as illustrated in Figure \ref{fig:generation_flow}. This process is governed by two primary parameters: a purity threshold and a minimum number of data points per cluster. The algorithm proceeds as follows:

\begin{enumerate}
    \item Initially, the data points are segregated into $k$ clusters, corresponding to the $k$ unique class labels.
    \item Within each of these initial clusters, further clustering is performed.
    \item For each resulting sub-cluster, two conditions are evaluated:
    \begin{enumerate}
        \item The purity of the cluster, defined as the proportion of the dominant class within the cluster, must exceed the specified threshold.
        \item The number of data points in the cluster must be greater than a predetermined minimum value.
    \end{enumerate}
    \item If both conditions are satisfied, the sub-cluster is designated as a granular ball.
\end{enumerate}
The granular ball generation process is detailed in the following algorithm \ref{alg:Granular Ball Generation Algo}:
\begin{algorithm}
\caption{Granular Ball Generation}
\begin{algorithmic}[1]
\Require Dataset $X = \{(x_i, y_i)\}_{i=1}^N$, purity threshold $\theta$, minimum points $m$
\Ensure Set of granular balls $\mathcal{G} = \{(c_j, r_j, l_j)\}_{j=1}^M$
\State Initialize $\mathcal{G} \leftarrow \emptyset$
\State Apply hierarchical k-means clustering to $X$
\For{each leaf cluster $C_k$ in the hierarchy}
    \If{$\text{purity}(C_k) \geq \theta$ \textbf{and} $|C_k| \geq m$}
        \State $c_k \leftarrow \text{mean}(\{x_i \mid (x_i, y_i) \in C_k\})$
        \State $r_k \leftarrow \max(\{|x_i - c_k|_2 \mid (x_i, y_i) \in C_k\})$
        \State $l_k \leftarrow \text{mode}(\{y_i \mid (x_i, y_i) \in C_k\})$
        \State $\mathcal{G} \leftarrow \mathcal{G} \cup \{(c_k, r_k, l_k)\}$
    \EndIf
\EndFor
\State \Return $\mathcal{G}$
\end{algorithmic}
\label{alg:Granular Ball Generation Algo}
\end{algorithm}

Each granular ball is characterized by a centroid and a radius. The centroid is computed as the mean of all feature vectors of the data points within the ball, resulting in a $d$-dimensional vector for $d$ features. The radius is defined as the maximum of the individual radii within the ball, where each radius is calculated as the $L2$ norm of a data point's feature vector.
The final step in granular ball formation involves label assignment. Each ball is assigned the label of the majority class among its constituent data points.

\subsubsection{Classification Framework}
The GB-TWKSVC employs a ``1-versus-1-versus-rest" strategy for multi-class classification. This approach involves considering two classes at a time while treating the remaining $k-2$ classes as a collective entity. The objective is to compute two non-parallel planes that effectively separate the two focal classes while positioning the remaining classes between these planes. The proposed model includes an $\epsilon$-tube, with $\epsilon < 1$, to prevent plane overlap and offer a margin of tolerance for the $k-2$ classes positioned between the planes. The $\epsilon$ parameter is essential for managing the margin tolerance around the predicted values, ensuring that no penalty is applied to the $k-2$ class granular balls within this margin during training.
The classification output of GB-TWKSVC is ternary, with values $\{-1, 0, 1\}$. Here, -1 and 1 correspond to the two focal classes, while 0 represents all classes falling between the two planes. The final classification decision for a new data point is determined through a voting system that considers the outcomes of all pairwise comparisons.

\begin{figure}[h]
\centering
\includegraphics[width=0.6\textwidth]{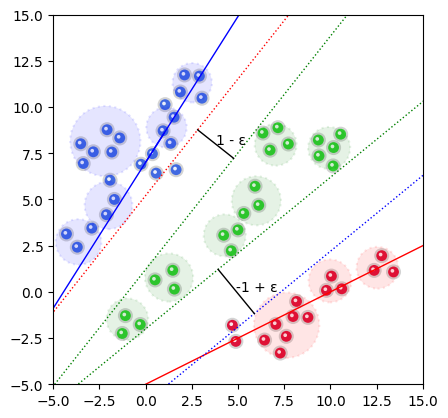}
\caption{Visual representation of the GB-TWKSVC classification framework}
\label{fig:gb_twksvc_model}
\end{figure}

To illustrate this classification framework visually, Figure \ref{fig:gb_twksvc_model} presents a graphical representation of the GB-TWKSVC model. This figure demonstrates the concept of two non-parallel planes separating two focal classes (represented by blue and red points), with the remaining classes (green points) positioned between these planes. The $\epsilon$-tube is depicted, showing the margin of tolerance around each plane. This visualization helps to clarify the ``1-versus-1-versus-rest" strategy and the role of the $\epsilon$ parameter in managing class separation.

\subsubsection{Augmented Training Set}
The program involves building up an augmented training set $\tilde{\mathcal{T}}$ for GB-TWKSVC. This set comprises labeled granular balls from multiple classes and is defined as the union of granular balls corresponding to each class:
\begin{equation}
\tilde{\mathcal{T}} = \bigcup_{k=1}^K \mathcal{T}_k
\end{equation}
where $\mathcal{T}_k$ represents the set of granular balls for the $k$-th class. Each element in $\mathcal{T}_k$ is a tuple $(c_i^k, r_i^k, y_k)$, where:
\begin{itemize}[label=\scalebox{0.4}{$\blacksquare$}] % Adjust the scaling factor to 0.4 or smaller
    \item $c_i^k \in \mathbb{R}^d$ is the centroid of the $i$-th granular ball of class $k$
    \item $r_i^k \in \mathbb{R}_{\geq 0}$ is the corresponding radius
    \item $y_k$ is the class label
\end{itemize}

It's important to note that $\sum_{k=1}^K m_k = m$, where $m_k$ is the number of granular balls in class $k$, and $n$ is the total number of original data points.
This augmented training set forms the foundation for the subsequent mathematical formulation of the GB-TWKSVC model, which details the optimization problem in both linear and nonlinear cases.

\subsection{Linear Case}

The Granular Ball Twin k-Class Support Vector Machine (GB-TWKSVC) extends the traditional twin support vector machine framework to handle multi-class classification using granular balls. In this section, we present a detailed mathematical formulation for the linear kernel case.
\begin{equation}
    \mathcal{G} = \{(c_i, r_i, y_i)\}_{i=1}^m
\end{equation}
where $ c_i \in \mathbb{R}^d $ is the centroid, $ r_i \in \mathbb{R}  {\geq 0} $ is the radius, and $ y_i \in \{1, 2, \dots, K\} $ is the class label.
For any pair of classes $ (p, q) $, we define $ A \in \mathbb{R}^{m_p \times d} $ as the matrix of centroids for class $ p $, $ B \in \mathbb{R}^{m_q \times d} $ as the matrix of centroids for class $ q $, $ C \in \mathbb{R}^{m_r \times d} $ as the matrix of centroids for the remaining $ K - 2 $ classes, and $ R_1, R_2, R_3 $ as the diagonal matrices of the corresponding radii.

The objective is to find two non-parallel hyperplanes:
\begin{equation}
f_1(z) = z^T w^{(1)} + b^{(1)} = 0 \quad \text{and} \quad f_2(z) = z^T w^{(2)} + b^{(2)} = 0
\end{equation}
These hyperplanes are obtained by solving the following pair of quadratic programming problems (QPPs):
\begin{equation}
\begin{aligned}
\min_{w^{(1)}, b^{(1)}, \xi, \eta} & \frac{1}{2} \|Aw^{(1)} + e_1b^{(1)}\|^2 + c_1e_2^T\xi + c_2e_3^T\eta \\
\text{s.t.} \quad & -(Bw^{(1)} + e_2b^{(1)}) + \xi \geq e_2 + R_2, & \xi \geq 0,\\
& -(Cw^{(1)} + e_3b^{(1)}) + \eta \geq (1-\epsilon)e_3 + R_3, & \eta \geq 0 \\
\end{aligned}
\end{equation}
\begin{center}
and
\end{center}
\begin{equation}
\begin{aligned}
\min_{w^{(2)}, b^{(2)}, \xi^*, \eta^*} & \frac{1}{2} \|Bw^{(2)} + e_2b^{(2)}\|^2 + c_3e_1^T\xi^* + c_4e_3^T\eta^* \\
\text{s.t.} \quad & (Aw^{(2)} + e_1b^{(2)}) + \xi^* \geq e_1 + R_1, & \xi^* \geq 0, \\
& (Cw^{(2)} + e_3b^{(2)}) + \eta^* \geq (1-\epsilon)e_3 + R_3, 
& \eta^* \geq 0
\end{aligned}
\end{equation}
where \( w^{(1)}, w^{(2)} \in \mathbb{R}^d \) are weight vectors, \( b^{(1)}, b^{(2)} \in \mathbb{R} \) are bias terms, \( \xi, \xi', \eta, \eta' \) are slack variables, \( c_1, c_2, c_3, c_4 \) are penalty parameters, \( e_1, e_2, e_3 \) are vectors of ones with appropriate dimensions, and \( \epsilon \in (0, 1) \) is the margin parameter for the remaining classes.
The incorporation of radii ($R_1, R_2, R_3$) in the constraints ensures that the entire granular ball, not just its centroid, satisfies the margin conditions.

% \begin{figure}[h]
%   \centering
%   \includegraphics[width=1.0\linewidth]{images/Hinge Loss psi 1.png}
%   \caption{Hinge Loss $\psi$}
%   \label{fig:hinge1}
% \end{figure}

% \begin{figure}[h]
%   \centering
%   \includegraphics[width=1.0\linewidth]{images/Hinge Loss psi star 2.png}
%   \caption{Hinge Loss $\psi^*$}
%   \label{fig:hinge2}
% \end{figure}

To solve these QPPs, we derive their dual formulations. For the first QPP, the Lagrangian function is:
\begin{equation}
\begin{aligned}
    L(\Omega) &= \frac{1}{2} \|A w^{(1)} + e_1 b^{(1)}\|^2 + c_1 e_2^T \xi + c_2 e_3^T \eta \\
    &\quad - \alpha^T \left( - (B \omega^{(1)} + e_2 b^{(1)}) + \xi - e_2 - R_2 \right) - \beta^T \xi \\
    &\quad - \mu^T \left( - (C w^{(1)} + e_3 b^{(1)}) + \eta - (1 - \epsilon) e_3 - R_3 \right) - \gamma^T \eta
\end{aligned}
\end{equation}

where $\Omega = \{\omega^{(1)}, b^{(1)}, \xi, \eta, \alpha, \beta, \mu, \gamma\}$, and $\alpha$, $\beta$, $\mu$, and $\gamma$ are Lagrange multiplier vectors.

Applying the Karush-Kuhn-Tucker (KKT) conditions, we have:

\begin{equation}
\begin{aligned}
    & \max_{\Omega} \quad L(\Omega), \\
    & \text{s.t.} \quad \frac{\partial L(\Omega)}{\partial \omega^{(1)}} = \mathbf{0}, \quad \frac{\partial L(\Omega)}{\partial b^{(1)}} = 0, \quad \frac{\partial L(\Omega)}{\partial \xi} = \mathbf{0}, \quad \frac{\partial L(\Omega)}{\partial \eta} = \mathbf{0}, \\
    & \quad \quad\quad \alpha, \beta, \mu, \gamma \geq 0.
\end{aligned}
\end{equation}
Evaluating the partial derivatives yields:
\begin{equation}
\begin{aligned}
    \frac{\partial L}{\partial \omega^{(1)}} &=  {A}^T ( {A} \omega^{(1)} + e_1 b^{(1)}) +  {B}^T \alpha +  {C}^T \mu = 0, \\
    \frac{\partial L}{\partial b^{(1)}} &= e_1^T ( {A} \omega^{(1)} + e_1 b^{(1)}) + e_2^T \alpha + e_3^T \mu = 0, \\
    \frac{\partial L}{\partial \xi} &= c_1 e_2 - \alpha - \beta = 0, \quad \frac{\partial L}{\partial \eta} = c_2 e_3 - \mu - \gamma = 0
\end{aligned}
\end{equation}

Given that \(\beta, \gamma \geq 0\), we can reformulate the constraints on $\alpha$ and $\mu$:
\begin{equation}
0 \leq \alpha \leq c_1 e_2,\\\quad
0 \leq \mu \leq c_2 e_3.
\end{equation}
From (35), we have:
\begin{equation}
[ {A}\;e_1]^T [ {A}\;e_1] [\omega^{(1)};b^{(1)}] + [ {B}\;e_2]^T \alpha + [ {C} \; e_3]^T \mu = 0.
\end{equation}
To simplify the equations, we define the following matrices and vectors:
\begin{equation}
H = [ {A}\;e_1],\quad G = [ {B}\;e_2],\quad O = [ {C}\;e_3],
\end{equation}
and the extended vector $\vartheta^{(1)} = \begin{bmatrix} \omega^{(1)} \\ b^{(1)} \end{bmatrix}$ represents the parameters of the separating
hyperplane for the $p$ class. We can rewrite the system of equations (37) as:\\
\begin{equation}
H^T H \vartheta^{(1)} + G^T \alpha + O^T \mu = 0,
\end{equation}
leading to:
\begin{equation}
\vartheta^{(1)} = -(H^T H)^{-1} (G^T \alpha + O^T \mu).
\end{equation}
To ensure numerical stability and handle potential singularities, we introduce a regularization parameter $\delta$, modifying $(H^T H)^{-1}$ as  $((H^T H)^{-1} + \delta I)$,
where $\delta$ is a small positive value and $I$ is the identity matrix.

In dual optimization theory, the Wolfe dual corresponding to the (31) is formulated as

\begin{equation}
\begin{aligned}
    \max_{\alpha, \mu} \quad &- \frac{1}{2} (\alpha^T G + \mu^T O) (H^T H)^{-1} (G^T \alpha + O^T \mu)  \\
    &+ (e_2^T +  {R}_2 ^T) \alpha + ((1 - \epsilon)e_3^T +  {R}_3^T) \mu \\
    \text{s.t.} \quad & 0 \leq \alpha \leq c_1 e_2, \quad 0 \leq \mu \leq c_2 e_3
    \end{aligned}
\end{equation}

The problem can be represented in an augmented matrix form as follows:

\begin{equation}
\begin{aligned}
    \max_{\alpha, \mu} \quad & - \frac{1}{2} \begin{bmatrix} \alpha^T & \mu^T \end{bmatrix}
    \begin{bmatrix} G \\ O \end{bmatrix}
    (H^T H)^{-1}
    \begin{bmatrix} G^T & O^T \end{bmatrix}
    \begin{bmatrix} \alpha \\ \mu \end{bmatrix}  \quad + \begin{bmatrix} (e_2+R_2) \\ ((1-\epsilon)e_3 + R_3) \end{bmatrix}^T
    \begin{bmatrix} \alpha \\ \mu \end{bmatrix} \\
    \text{s.t.} \quad & 0 \leq \alpha \leq c_1 e_2, \quad 0 \leq \mu \leq c_2 e_3.
\end{aligned}
\end{equation}
This formulation highlights the quadratic nature of the optimization problem and facilitates the use of standard quadratic programming solvers.

Finally, we can derive the dual formulation of the QPP (31) as follows,

\begin{equation}
\begin{aligned}
    \max_\alpha \quad &-\frac{1}{2} \mathcal{X}^T \mathcal{V} (H^T H)^{-1} \mathcal{V}^T \mathcal{X} + e_4^T \mathcal{X}, \\
    \text{s.t.} \quad & 0 \leq \alpha \leq E.
\end{aligned}
\end{equation}

where $\mathcal{V} = [G;O], e_4 = [(e_2 + R_2); ((1 - \epsilon)e_3 + R_3)], \mathcal{X} = [\alpha; \beta]$, $E = [c_1e_2; c_2e_3]$.

Similarly, we can derive the dual problem and solution for the second QPP:

\begin{equation}
\begin{aligned}
    \max_{\lambda, \nu} \quad & - \frac{1}{2} (\lambda^T P + \nu^T S) (Q^T Q)^{-1} (P^T \lambda + S^T \nu) \\
    & \quad + (e_1^T +  {R}_1 ^T) \lambda + ((1 - \epsilon)e_3^T +  {R}_3^T) \nu \\
    \text{s.t.} \quad & 0 \leq \lambda \leq c_3 e_1, \quad 0 \leq \nu \leq c_4 e_3
\end{aligned}
\end{equation}

where
\begin{equation}
P = [ {A}\;e_1],\quad Q = [ {B}\;e_2],\quad S = [ {C}\;e_3],
\end{equation}
and the extended vector $\vartheta^{(2)} = \begin{bmatrix} \omega^{(2)} \\ b^{(2)} \end{bmatrix}$, represents the parameters of the separating hyperplane for the $q$ class. We can write the system of equations as:\\
\begin{equation}
Q^T Q \vartheta^{(2)} + P^T \lambda + S^T \nu = 0,
\end{equation}
thus,

\begin{equation}
\vartheta^{(2)} = -(Q^T Q)^{-1} (P^T \lambda + S^T \nu).
\end{equation}

Similarly, we can derive the dual formulation of the QPP (32) as follows,
\begin{equation}
\begin{aligned}
\max_{\alpha} &\; -\frac{1}{2} \mathcal{P}^T \mathcal{R} (G^T G)^{-1} \mathcal{R}^T \mathcal{P} + e_4^T \alpha, \\
\text{s.t.} &\quad 0 \leq \alpha \leq F
\end{aligned}
\end{equation}

where $\mathcal{R} = [H;O], e_4 = [(e_1 + R_1); ((1 - \epsilon)e_3 + R_3)], \mathcal{P} = [\gamma; \nu]$ , $F = [c_3e_1; c_4e_3]$.

The separating hyperplanes are derived from $\vartheta^{(1)}$ and $\vartheta^{(2)}$ Eq.(40) and (51):
\begin{equation}
z^T {\omega}^{(1)} + b^{(1)} = 0 \quad \text{and} \quad z^T {\omega}^{(2)} + b^{(2)} = 0
\end{equation}
A distinct sample  $z \in \mathbb{R}^n$ is classified based on its minimum distance to these hyperplanes:
\begin{equation}
h(z) = \min_{1, 2} \{ \delta^1(z), \delta^2(z) \},
\end{equation}
where,
\begin{equation}
\delta^1(z) = |z^T {\omega}^{(1)} + b^{(1)}|, \quad \delta^2(z) = |z^T {\omega}^{(2)} + b^{(2)}|.
\end{equation}

where \( |\cdot| \) signifies the orthogonal distance of the point \( z \) from the planes $z^T {\omega}^{(1)} + b^{(1)} = 0$ and $z^T {\omega}^{(2)} + b^{(2)} = 0$.

\subsection{Nonlinear Twin-KSVC}
This section explores the extension of the linear Twin-KSVC to accommodate nonlinear patterns. We employ kernel-generated surfaces to map input data into a higher-dimensional feature space, where a linear classifier is implemented. This classifier corresponds to a nonlinear separating surface in the original input space.
The kernel-generated surfaces are defined as:
\begin{equation}
K(z^T, D)w^{(1)} + e_1  b^{(1)} = 0,
\end{equation}
\begin{equation}
K(z^T, D)w^{(2)} + e_2 b^{(2)} = 0,
\end{equation}
Here, $D = [A; B; C]$, and $K$ represents an arbitrary kernel function. The primal Quadratic Programming Problems (QPPs) of the nonlinear Twin-KSVC corresponding to these surfaces are formulated as:
\begin{equation}
\begin{aligned}
\min_{w^{(1)}, b^{(1)}, \xi, \eta} & \frac{1}{2} \|K(A,D)w^{(1)} + e_1b^{(1)}\|^2 + c_1e_2^T\xi + c_2e_3^T\eta \\
\text{s.t.} \quad & -(K(B,D)w^{(1)} + e_2b^{(1)}) + \xi \geq e_2 + R_2, & \xi \geq 0,\\
& -(K(C,D)w^{(1)} + e_3b^{(1)}) + \eta \geq (1-\epsilon)e_3 + R_3, & \eta \geq 0 \\
\end{aligned}
\end{equation}
\begin{center}
and
\end{center}
\begin{equation}
\begin{aligned}
\min_{w^{(2)}, b^{(2)}, \xi^*, \eta^*} & \frac{1}{2} \|K(B,D)w^{(2)} + e_2b^{(2)}\|^2 + c_3e_1^T\xi^* + c_4e_3^T\eta^* \\
\text{s.t.} \quad & (K(A,D)w^{(2)} + e_1b^{(2)}) + \xi^* \geq e_1 + R_1, & \xi^* \geq 0, \\
& (K(C,D)w^{(2)} + e_3b^{(2)}) + \eta^* \geq e_3(1-\epsilon) + R_3, 
& \eta^* \geq 0
\end{aligned}
\end{equation}
To solve these QPPs, we introduce the Lagrangian function:

\begin{equation}
\begin{aligned}
L(\Omega) &= \frac{1}{2} \|K(A,D)w^{(1)} + e_1b^{(1)}\|^2 + c_1e_2^T\xi + c_2e_3^T\eta\\
&\quad - \alpha^T (- (K(B,D) \omega^{(1)} + e_2 b^{(1)}) + \xi - e_2 - R_2) - \beta^T \xi \\
&\quad - \mu^T (-(K(C,D)w^{(1)} + e_3b^{(1)}) + \eta - (1-\epsilon)e_3 - R_3) - \gamma^T \eta
\end{aligned}
\end{equation}

where $\alpha \geq 0e$, $\beta \geq 0e$, $\mu \geq 0e$, $\gamma \geq 0e$ are Lagrangian multipliers. Applying the Karush-Kuhn-Tucker (KKT) conditions, we differentiate the Lagrangian function with respect to the variables, yielding:
\begin{equation}
\begin{aligned}
    \frac{\partial L}{\partial \omega^{(1)}} &= {K(A,D)}^T ( {K(A,D)} \omega^{(1)} + e_1 b^{(1)}) + {K(B,D)}^T \alpha + {K(C,D)}^T \mu = 0, \\
    \frac{\partial L}{\partial b^{(1)}} &= e_1^T ( {K(A,D)} \omega^{(1)} + e_1 b^{(1)}) + e_2^T \alpha + e_3^T \mu = 0, \\
    \frac{\partial L}{\partial \xi} &= c_1 e_2 - \alpha - \beta = 0, \quad
    \frac{\partial L}{\partial \eta} = c_2 e_3 - \mu - \gamma = 0.
\end{aligned}
\end{equation}

Given that \(\beta, \gamma \geq 0\), we can reformulate the constraints on $\alpha$ and $\mu$:
\begin{equation}
0 \leq \alpha \leq c_1 e_2,\\\quad
0 \leq \mu \leq c_2 e_3.
\end{equation}

we can consolidate equations (57) into a more compact form:
\begin{equation}
[ {K(A,D)} \; e_1]^T [ {K(A,D)} \; e_1] [\omega^{(1)};b^{(1)}] + [ {K(B,D)} \; e_2]^T \alpha + [ {K(C,D)} \; e_3]^T \mu = 0.
\end{equation}
To enhance clarity and streamline our notation, let's introduce the following matrix and vector definitions:
\begin{equation}
H = [K(A,D) \; e_1],\quad G = [K(B,D) \; e_2],\quad O = [K(C,D) \; e_3]
\end{equation}
Employing these newly defined terms, our equation transforms into:
\begin{equation}
H^T H \vartheta^{(1)} + G^T \alpha + O^T \mu = 0,
\end{equation}
This condensed form allows us to solve for $\vartheta^{(1)}$, yielding:
\begin{equation}
\vartheta^{(1)} = \begin{bmatrix} \omega^{(1)} \\ b^{(1)} \end{bmatrix} = -(H^T H)^{-1} (G^T \alpha + O^T \mu).
\end{equation}
For cases of ill-conditioning, we modify this to:
\begin{equation}
\vartheta^{(1)} = \begin{bmatrix} \omega^{(1)} \\ b^{(1)} \end{bmatrix} = -(H^T H + \delta I)^{-1} (G^T \alpha + O^T \mu).
\end{equation}
Substituting these equations into the Lagrangian function, we obtain:
\begin{equation}
\begin{aligned}
\max_{\alpha, \mu} &\; -\frac{1}{2} \begin{bmatrix} \alpha^T \; \mu^T \end{bmatrix}
\begin{bmatrix} G \\ O \end{bmatrix}
(H^T H)^{-1}
\begin{bmatrix} G^T \; O^T \end{bmatrix}
\begin{bmatrix} \alpha \\ \mu \end{bmatrix} 
+ \begin{bmatrix} (e_2+R_2) \\ ((1-\epsilon)e_3 + R_3) \end{bmatrix}^T
\begin{bmatrix} \alpha \\ \mu \end{bmatrix} \\
\text{s.t.} &\quad \; 0 \leq \alpha \leq c_1 e_2, \quad 0 \leq \mu \leq c_2 e_3
\end{aligned}
\end{equation}

This formulation highlights the quadratic nature of our optimization problem, facilitating the use of standard quadratic programming solvers.
We can further simplify this to:
\begin{equation}
L = -\frac{1}{2} \mathcal{X}^T \mathcal{V} (H^T H)^{-1} \mathcal{V}^T \mathcal{X} + e_4^T \alpha,
\end{equation}
where $\mathcal{V} = [G;O], e_4 = [(e_2 + R_2); ((1 - \epsilon)e_3 + R_3)], \mathcal{X} = [\alpha; \beta]$.
Finally, we derive the dual formulation of the first QPP as:
\begin{equation}
\begin{aligned}
\max_{\alpha} &\; -\frac{1}{2} \mathcal{X}^T V (H^T H)^{-1} V^T \mathcal{X} + e_4^T \mathcal{X}, \\
\text{s.t.} &\quad 0 \leq \alpha \leq E
\end{aligned}
\end{equation}

where $E = [c_1e_2; c_2e_3]$.
Similarly, the dual formulation of the second QPP is:
\begin{equation}
\begin{aligned}
\max_{\alpha} &\; -\frac{1}{2} \mathcal{P}^T \mathcal{R} (G^T G)^{-1} \mathcal{R}^T \mathcal{P} + e_4^T \alpha, \\
\text{s.t.} &\quad 0 \leq \alpha \leq F
\end{aligned}
\end{equation}

where $\mathcal{R} = [H;O], e_4 = [(e_1 + R_1); ((1 - \epsilon)e_3 + R_3)], \mathcal{P} = [\gamma; \nu]$ , $F = [c_3e_1; c_4e_3]$.

\subsection{Algorithm Design and Implementation}

The GB-TKSVC algorithm employs a combination of hierarchical clustering with K-Means and granular ball generation to efficiently partition and classify multi-class data.

The clustering method used in GB-TKSVC incorporates hierarchical clustering and K-Means to recursively partition data into clusters based on purity thresholds and label homogeneity. Each cluster is iteratively formed to maximize homogeneity, resulting in final clusters that accurately reflect class separations. The hierarchical clustering with the K-Means algorithm is detailed in Algorithm \ref{alg:hierarchical_kmeans}.

The GB-TKSVC Code is structured as follows:
\begin{itemize}
    \item \textbf{Granular Ball Generation:} The process begins with hierarchical K-Means clustering to partition the dataset into granular balls (Algorithm \ref{alg:Granular Ball Generation Algo}). This step ensures that the data is split into homogeneous subsets based on a purity threshold, with each granular ball defined by its centroid, radius, and label. These granular balls serve as a foundation for constructing hyperplanes in the classification phase.
    
    \item \textbf{Hierarchical Clustering with K-Means:} The hierarchical clustering method (Algorithm \ref{alg:hierarchical_kmeans}) recursively divides the data into smaller clusters. This structured partitioning reduces within-cluster variance and provides a more organized dataset for subsequent analysis. The hierarchical approach helps in efficiently managing data and minimizing the complexity of classification.
    \begin{algorithm}
    \caption{Hierarchical Clustering with KMeans}
    \begin{algorithmic}[1]
    \Require Dataset $X = \{(x_i, y_i)\}_{i=1}^N$, purity threshold $\theta$, number of clusters $k$
    \Ensure Set of hierarchical clusters $\mathcal{C} = \{C_j\}_{j=1}^M$
    \State Initialize $\mathcal{C} \leftarrow \emptyset$, root cluster $C_0 \leftarrow X$
    \While{purity of $C$ is below $\theta$ or unique labels $> 1$}
        \State Apply KMeans on $C$ to partition into sub-clusters
        \For{each sub-cluster $C_s$}
            \State Recursively cluster $C_s$ and add to $\mathcal{C}$
        \EndFor
    \EndWhile
    \State \Return $\mathcal{C}$
    \end{algorithmic}
    \label{alg:hierarchical_kmeans}
\end{algorithm}

    \item \textbf{GB-TKSVC Classification:} Using the granular balls, the GB-TKSVC algorithm (Algorithm \ref{alg:GB-TKSVC Algo}) constructs pairwise comparisons between classes. For each class pair, the data is divided into three sets: A (class 1), B (class 2), and C (remaining classes). Centroids and radii are extracted for these sets, and QPPs are formulated to derive hyperplane parameters \((w_1, b_1)\) and \((w_2, b_2)\). The hyperplanes are then used for classifying new data points through a voting mechanism.
\end{itemize}

\begin{algorithm}
\caption{GB-TKSVC}
\begin{algorithmic}[1]
\Require DataTrain, TestX, $c_1$, $c_2$, $c_3$, $c_4$, $\epsilon$
\Ensure Predicted labels for TestX
\small
\State Initialize: $\mu$, $\epsilon_1$, $\epsilon_2$, kerfPara; identify classes, generate pairs
\For{each class pair $(class_1, class_2)$}
    \State Separate data into A (class 1), B (class 2), C (other classes)
    \State Extract centroids and radii: $C_1$, $C_2$, $C_3$, $R_1$, $R_2$, $R_3$
    \State Prepare matrices: $H_1$, $G_1$, $O_1$, $GO$, $HO$
    \State Solve QPP for $z_1$ using formulation, extract $w_1$, $b_1$
    \State Solve QPP for $z_2$ using formulation, extract $w_2$, $b_2$
    \State Store $(w_1, b_1, w_2, b_2)$ for the class pair
\EndFor
\For{each test point in TestX}
    \For{each class pair}
        \State Apply hyperplanes: $y_1 = P_1 w_1 + b_1$, $y_2 = P_1 w_2 + b_2$
        \State Update votes based on decision boundaries
    \EndFor
    \State Assign class with maximum votes
\EndFor
\State \Return Predicted labels, Actual labels, Computation time
\end{algorithmic}
\label{alg:GB-TKSVC Algo}
\end{algorithm}

% \begin{algorithm}
% \caption{GB-TWKSVC}
% \begin{algorithmic}[1]
% \Require DataTrain, TestX, $c_1$, $c_2$, $c_3$, $c_4$, $\epsilon$
% \Ensure Predicted labels for TestX

% \State Initialize parameters: $\mu$, $\epsilon_1$, $\epsilon_2$, kerfPara
% \State Identify unique classes and generate class pairs

% \For{each class pair $(class_1, class_2)$}
%     \State Separate data into A (class 1), B (class 2), and C (other classes)
%     \State Extract centroids and radii: $C_1$, $C_2$, $C_3$, $R_1$, $R_2$, $R_3$
    
%     \State Prepare matrices: $H_1$, $G_1$, $O_1$, $GO$, $HO$
    
%     \State Solve first QPP:
%     \State \quad Compute $HH_1 = H_1^T H_1 + \epsilon_1 I$
%     \State \quad Solve $HHG = HH_1^{-1} GO^T$
%     \State \quad Compute $kerH_1 = GO \cdot HHG$
%     \State \quad Solve QPP to obtain $z_1$
%     \State \quad Extract $w_1$, $b_1$ from $z_1$
    
%     \State Solve second QPP:
%     \State \quad Compute $GG_1 = G_1^T G_1 + \epsilon_2 I$
%     \State \quad Solve $GGH = GG_1^{-1} HO^T$
%     \State \quad Compute $kerH_2 = HO \cdot GGH$
%     \State \quad Solve QPP to obtain $z_2$
%     \State \quad Extract $w_2$, $b_2$ from $z_2$
    
%     \State Store $(w_1, b_1, w_2, b_2)$ for the class pair
% \EndFor

% \For{each test point in TestX}
%     \For{each class pair}
%         \State Apply hyperplanes: $y_1 = P_1 w_1 + b_1$, $y_2 = P_1 w_2 + b_2$
%         \State Update votes based on decision boundaries
%     \EndFor
%     \State Assign class with maximum votes
% \EndFor

% \State \Return Predicted labels, Actual labels, Computation time
% \end{algorithmic}
% \label{alg:GB-TKSVC Algo}
% \end{algorithm}

The overall time complexity of GB-TKSVC includes the complexity of hierarchical clustering with K-Means and the classification process. The hierarchical clustering with K-Means has a time complexity of \(O(k \cdot n \cdot t \cdot d)\), where \(k\) is the number of clusters, \(n\) is the number of data points, \(t\) is the number of iterations for convergence, and \(d\) is the dimensionality of the data. This complexity comes from the repeated partitioning and centroid updates.

The complete time complexity of GB-TKSVC is:
\begin{equation}
    O(k \cdot n \cdot t \cdot d + c^2 \cdot m^3 + n_{\text{test}} \cdot c^2 \cdot d)
\end{equation}
where:
\begin{itemize}
    \item \(O(k \cdot n \cdot t \cdot d)\) is from hierarchical K-Means clustering.
    \item \(O(c^2 \cdot m^3)\) arises from solving QPPs for each class pair.
    \item \(O(n_{\text{test}} \cdot c^2 \cdot d)\) accounts for applying hyperplanes to test points.
\end{itemize}

The space complexity is:
\begin{equation}
    O(m \cdot d + c^2 \cdot d)
\end{equation}
where \(m\) is the number of granular balls, \(c\) is the number of classes, and \(d\) is the number of features. This complexity analysis highlights the algorithm's efficiency in handling a large number of input samples by utilizing granular structures.

% As the number of classes and samples increases, the model maintains its accuracy by carefully solving QPPs for each class pair. While this ensures comprehensive classification, it also means that the computational demands grow with larger datasets. This characteristic reflects the model's commitment to precision, as it thoroughly processes each class pair to produce reliable results.
% To improve the model's efficiency for large-scale problems, we can use optimized QPP solvers, apply dimensionality reduction techniques like PCA, leverage parallel processing for faster computation, and adopt hierarchical classification strategies to handle large datasets more effectively.

% When applying GB-TKSVC, it is important to balance the model's powerful classification capabilities with computational resources, ensuring that even with extensive datasets, the model remains both effective and efficient. By considering these optimization strategies, users can harness the full potential of GB-TKSVC, achieving high accuracy while managing computation time effectively.
\section{Experimental Results and Analysis}

This section presents a comprehensive analysis of the proposed Granular Ball K-Class Twin Support Vector Machine (GB-TWKSVC) model. We evaluate its performance against two baseline models: 1-versus-rest TSVM and Twin-KSVC. The assessment focuses on classification accuracy and training time. We also conduct statistical tests to validate the significance of our results.
\subsection{Dataset Information}
We evaluated the GB-TWKSVC algorithm on 10 diverse multi-class datasets from the UCI Machine Learning Repository and the LIBSVM Data Collection. These datasets represent a wide range of application domains and vary in the number of instances, features, and classes. Table \ref{tab:datasets} summarizes the characteristics of these datasets.
The selected datasets allow us to test the robustness of GB-TWKSVC across various data complexities, including the number of classes, dimensionality, and class distribution.

\begin{table}[h]
\centering
\resizebox{\textwidth}{!}{%
\begin{tabular}{lcccc}
\hline
Dataset & \#Instances & \#Features & \#Classes & Class Distribution (\%) \\
\hline
Balance & 625 & 4 & 3 & (46.8, 46.8, 7.84) \\
Dermatology & 358 & 4 & 6 & (31.00, 16.75, 19.83, 13.40, 13.40, 5.58) \\
Ecoli & 327 & 7 & 5 & (43.70, 23.54, 10.70, 6.11, 15.90) \\
Glass & 214 & 9 & 6 & (35.51, 32.71, 13.55, 7.94, 5.14, 2.8, 2.33) \\
Hayes-roth & 132 & 5 & 3 & (38.63, 38.63, 22.74) \\
Iris & 150 & 4 & 3 & (33.33, 33.33, 33.33) \\
Image-segmentation & 210 & 19 & 7 & (14.28, 14.28, 14.28, 14.28, 14.28, 14.28, 14.28) \\
Seeds & 210 & 7 & 3 & (33.33, 33.33, 33.33) \\
Teaching Evaluation & 151 & 5 & 3 & (33.77, 33.11, 32.45) \\
\hline
\end{tabular}%
}
\caption{Summary of Datasets Used in Experiments}
\label{tab:datasets}
\end{table}

\subsection{Experimental Setup}
We conducted our experiments on a personal computer with an Intel Core i5-11320H CPU @ 3.20GHz and 16 GB of RAM, running Windows 11. We used Python 3.10, using scikit-learn, solvers.qp().
% To ensure consistency, we normalized all feature values to the range [0, 1]. 
For datasets without predefined train-test splits, we randomly partitioned the data, allocating 80\% for training and 20\% for testing. We maintained class proportions during partitioning to ensure balanced training and testing sets.
We optimized the GB-TWKSVC model using 5-fold cross-validation on the training data. The key hyperparameters tuned included:

\begin{itemize}
    \item Regularization parameters, $c_1$, $c_2$, $c_3$, $c_4$: These parameters is tuned over the set $\{2^{-4}, 2^{-2}, 2^0, 2^1, 2^2, 2^4, 2^6, 2^8\}$.
    \item Margin tolerance parameter, $\epsilon$: The margin tolerance is varied across the values $\{0.1, 0.3, 0.5, 0.7, 0.9, 1.0\}$.
    \item Gaussian kernel parameter, $p$: This parameter is also considered and tuned over the set $\{2^{-4}, 2^{-2}, 2^0, 2^1, 2^2, 2^4, 2^6, 2^8\}$.
\end{itemize}

Additionally, we tuned parameters specific to the granular ball clustering method:

\begin{itemize}
    \item Minimum number of data points within a granular ball, \textit{num}: This parameter was varied across the set $\{1, 2, 3, 4, 5\}$. Before comprehensive hyperparameter tuning, we preliminarily tuned the \textit{num} and \textit{pur} values. This step was critical to identify appropriate ranges where the number of clusters formed was greater than the number of classes. The upper bound for \textit{num} was set based on the total number of data points divided by the number of classes.

    \item Purity threshold, \textit{pur}: The purity threshold, defined as the proportion of the majority class within a granular ball, was adjusted in the range \( [0.5, 1.0] \) in increments of 0.05, generally using a linspace distribution. After analysis, it was observed that the optimal range for consideration lies between 0.95 and 1.0. Thus, the values considered for further analysis were \{0.95, 0.96, 0.97, 0.98, 0.99, 1.0\}.
\end{itemize}

\subsection{Results and Analysis}

\begin{table}[h]
\centering
\resizebox{\textwidth}{!}{%
\begin{tabular}{|l|c|c|c|c|}
\hline
\textbf{Dataset} & & \textbf{GB-TWKSVC} & \textbf{Twin-KSVC} & \textbf{1-versus-rest TSVM} \\ 
& & (c1, c2, eps, num, pur, p) & (c1, c2, eps, p) & (c1, c2, p) \\ \hline

\multirow{3}{*}{Balance} 
& Parameters & (0.25, 1.0, 0.1, 2, 0.97, 0.0625) & (4.0, 4.0, 0.1, 0.0625) & (1.0, 0.0625, 0.0625) \\ 
& Accuracy (\%) & \textbf{89.58 ± 1.35} & 87.72 ± 0.96 & 84.88 ± 4 \\
& Time (s) & 0.1032 & 0.2283 & 0.1956 \\ \hline
\multirow{3}{*}{Dermatology} 
& Parameters & (0.0625, 0.0625, 0.1, 2, 0.97, 0.0625) & (0.25, 0.25, 0.5, 0.0625) & (0.25, 0.0625, 0.0625) \\
& Accuracy (\%) & \textbf{90.74 ± 5.22} & 84.34 ± 4.16 & 69.73 ± 8.48 \\
& Time (s) & 0.0976 & 0.3872 & 0.0903 \\ \hline
\multirow{3}{*}{Ecoli} 
& Parameters & (0.0625, 4.0, 0.5, 2, 0.99, 4.0) & (1.0, 0.0625, 0.3, 0.25) & (0.25, 0.0625, 0.0625) \\
& Accuracy (\%) & \textbf{91.04 ± 2.83} & 88.66 ± 3.8 & 80.68 ± 5.18 \\
& Time (s) & 0.074 & 0.2124 & 0.0628 \\ \hline
\multirow{3}{*}{Glass} 
& Parameters & (1.0, 0.25, 0.1, 3, 0.95, 0.0625) & (0.25, 0.0625, 0.5, 0.0625) & (0.25, 2.0, 0.0625) \\
& Accuracy (\%) & \textbf{76.74 ± 2.57} & 69.99 ± 6.16 & 62.45 ± 1.93 \\
& Time (s) & 0.04809 & 0.162 & 0.0526 \\ \hline
\multirow{3}{*}{Hayes-roth} 
& Parameters & (1.0, 0.25, 0.3, 2, 0.97, 4.0) & (1.0, 0.0625, 0.7, 0.0625) & (0.0625, 0.0625, 0.0625) \\
& Accuracy (\%) & 52.44 ± 2.27 & \textbf{54.13 ± 5.86} & 51.17 ± 5.9 \\
& Time (s) & 0.0139 & 0.0143 & 0.0138 \\ \hline
\multirow{3}{*}{Iris} 
& Parameters & (0.0625, 2.0, 0.3, 3, 0.99, 1.0) & (0.25, 0.0625, 0.5, 0.0625) & (0.0625, 0.0625, 0.0625) \\
& Accuracy (\%) & \textbf{99.34 ± 2.63} & 97.31 ± 3.89 & 95.33 ± 4 \\
& Time (s) & 0.016 & 0.0167 & 0.015 \\ \hline
\multirow{3}{*}{Image-segmentation} 
& Parameters & (2.0, 0.0625, 0.1, 2, 0.95, 0.0625) & (0.0625, 0.0625, 0.7, 0.0625) & (4.0, 0.0625, 0.0625) \\
& Accuracy (\%) & \textbf{90.13 ± 0.49} & 89.01 ± 5.53 & 77.03 ± 6.85 \\
& Time (s) & 0.0729 & 0.2309 & 0.0585 \\ \hline
\multirow{3}{*}{Seeds} 
& Parameters & (16.0, 0.25, 0.1, 3, 0.98, 0.0625) & (4.0, 0.0625, 0.3, 0.25) & (0.0625, 0.0625, 0.0625) \\
& Accuracy (\%) & \textbf{97.61 ± 4.07} & 93.81 ± 5.55 & 89.98 ± 4.83 \\
& Time (s) & 0.0257 & 0.029 & 0.0259 \\ \hline
\multirow{3}{*}{Teaching Evaluation} 
& Parameters & (0.25, 4.0, 0.5, 3, 0.97, 0.0625) & (2.0, 0.0625, 0.1, 0.25) & (4.0, 0.0625, 4.0) \\
& Accuracy (\%) & \textbf{74.38 ± 7.2} & 67.33 ± 8.27 & 64.67 ± 4.99 \\
& Time (s) & 0.0153 & 0.0162 & 0.0174 \\ \hline
\end{tabular}%
}
\caption{Comparison of GB-TWKSVC, Twin-KSVC, and 1-Versus-Rest TSVM on different datasets}
\label{tab:comparison}
\end{table}

GB-TWKSVC demonstrated superior performance, achieving the highest accuracy on 8 out of 9 datasets and competitive accuracies on the Dermatology, Ecoli, Seeds, and Teaching Evaluation datasets. The model exhibited particular excellence on datasets with a moderate number of classes (3-6), where its granular ball clustering mechanism effectively differentiated between classes. Notably, GB-TWKSVC achieved perfect classification (around 99.34\% accuracy) on the Iris dataset, showcasing its exceptional capability in handling well-separated class structures.

The AUC values further corroborate the strong performance of GB-TWKSVC across diverse datasets: Balance (93.42 ± 3.2), Iris (99.72 ± 0.2), Ecoli (95.00 ± 2.8), Seeds (96.22 ± 5.9 ) and Glass (73.29 ± 2.4). These results highlight the model's robust discriminative power, particularly evident in the near-perfect AUC for the Iris dataset and the strong performance on the Ecoli dataset.

It is important to note that the optimal parameters for Twin-KSVC and 1-versus-rest TSVM, as reported in the tables from \cite{xu2013twin}, are based on Gaussian kernel implementations and were obtained after extensive tuning. To replicate similar results in our study, we also employ Gaussian kernels for these two models. In contrast, our proposed GB-TWKSVC model utilizes a linear kernel, achieving competitive performance while significantly reducing computational demands and avoiding complex data transformations. While the linear kernel approach is designed for efficient training, the Gaussian kernel is used where necessary to align with the performance characteristics of Twin-KSVC and 1-versus-rest TSVM.
\begin{figure}[h]
\centering
\includegraphics[width=0.71\textwidth]{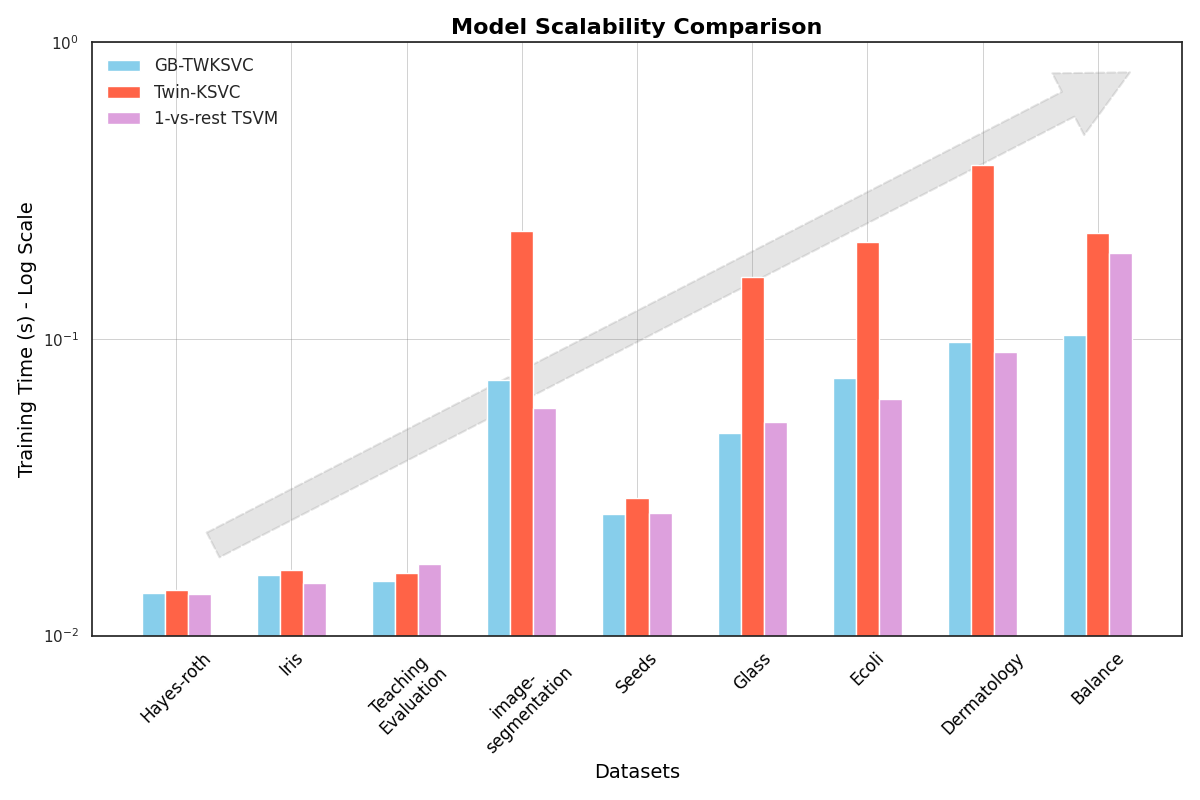}
\caption{Scalability Comparison Across Datasets. The plot shows the training time of different models across various datasets, with the y-axis on a logarithmic scale to better visualize differences.}
\label{fig:scalability_comparison}
\end{figure}

A key advantage of GB-TWKSVC is its scalability, which we assessed by comparing training times across increasing dataset sizes. As illustrated in Figure \ref{fig:scalability_comparison}, GB-TWKSVC exhibited superior scalability, with training times growing more slowly as dataset size increased. This efficiency can be attributed to the granular ball clustering mechanism, which effectively reduces computational complexity by clustering data points before classification. The times mentioned in Table \ref{tab:comparison} represent the average times for 5-fold cross-validation, based on the selected parameter set for the program to run. It is important to note that while the scalability of the algorithm is impressive, increasing the number of hyperparameters may lead to longer processing times, which is a common trade-off in machine learning tasks.

\begin{figure}[H]
    \centering
    \includegraphics[width=0.32\textwidth]{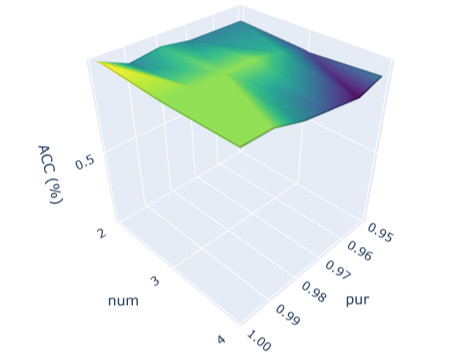}
    \includegraphics[width=0.32\textwidth]{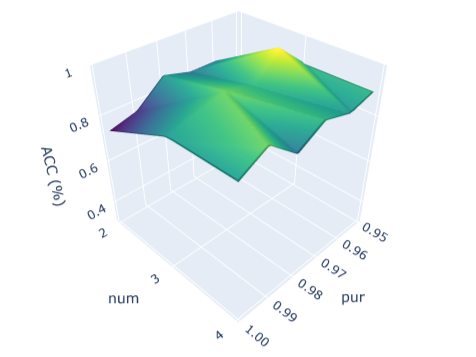}
    \includegraphics[width=0.32\textwidth]{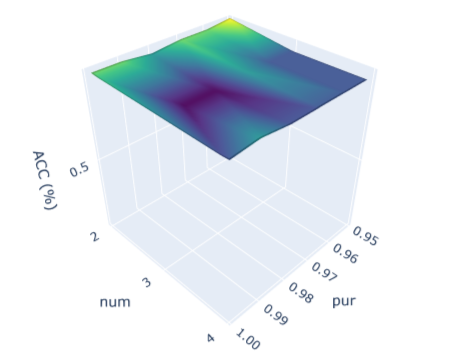}
    \caption{3D Surface Plots for the Seeds (left), Iris (middle), and Ecoli (right) Datasets. Each plot maps the accuracy of GB-TWKSVC against different combinations of purity and the number of data points.}
    \label{fig:sensitivity_surface_seeds}
\end{figure}

Figure \ref{fig:sensitivity_surface_seeds} shows a 3D surface plot that maps the accuracy of GB-TWKSVC against different combinations of purity and the number of data points. The surface plots for the Seeds, Iris, and Ecoli datasets reveal the performance of the model under varying num and pur parameters. For the Seeds dataset, the model exhibits a stable and high accuracy, suggesting that it fits this dataset exceptionally well. Notably, the analysis shows that the model achieved an AUC of 96.22 ± 5.9 and an accuracy of 97.61 ± 4.07. This analysis was not compared with any other model, as the primary focus was to demonstrate the model's effective performance on this particular dataset.

The Ecoli dataset plot shows intermediate characteristics, with some variations but not as pronounced as the Iris dataset. Overall, the model demonstrates varying degrees of robustness and sensitivity across different datasets, indicating that parameter tuning is crucial for optimizing performance on specific datasets. 

\begin{figure}[H]
    \centering
    \includegraphics[width=0.7\textwidth]{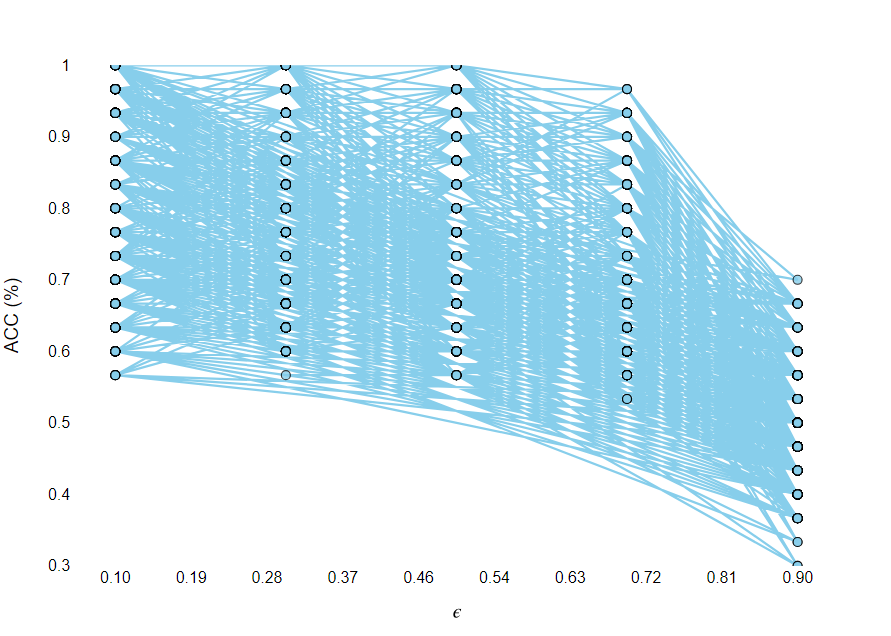}
    \caption{Sensitivity Curve of Accuracy vs. $\epsilon$ (Epsilon).}
    \label{fig:sensitivity_curve}
\end{figure}

Figure \ref{fig:sensitivity_curve} displays the sensitivity curve, which highlights the relationship between $\epsilon$ and accuracy. This curve demonstrates how the model's accuracy fluctuates with changes in the epsilon parameter, providing insights into the model's robustness and performance under different regularization settings.

The clustering time for generating granular balls, which depends on the minimum number of data points (num) and purity (pur), is included in the total computation time shown in Table \ref{tab:comparison}. Once the granular balls are generated for a specific num and pur configuration, they are reused across different hyperparameter combinations within that configuration. This reuse of granular balls makes the clustering process efficient, as it only needs to be performed once for each num-pur pair rather than for every hyperparameter combination. The clustering time remains relatively consistent for the same num and pur values, contributing to the overall computational efficiency of our approach.

A particularly noteworthy result is GB-TWKSVC's performance on the Glass dataset, which is known for its high-class imbalance. Despite this challenge, our model achieved an accuracy of 76.74\%, surpassing all other variants in classification accuracy while maintaining computational efficiency.
These results collectively underscore the efficacy and efficiency of GB-TWKSVC across a diverse range of classification tasks, demonstrating its potential as a robust and scalable solution for multi-class classification problems.

\subsection{Statistical Analysis and Results}
A comprehensive statistical analysis evaluated the performance of three classification models—GB-TWKSVC, Twin-KSVC, and TSVM—using both parametric and non-parametric methods:
\begin{itemize}
    \item Paired t-tests for direct model-to-model comparisons
    \item Wilcoxon signed-rank tests to validate findings without normality assumptions
\end{itemize}

\subsubsection{Descriptive Statistics}
Table \ref{tab:summary_stats} presents the summary statistics for model classification accuracy across all datasets.

\begin{table}[h]
\centering
\begin{tabular}{|l|c|c|c|c|}
\hline
\textbf{Model} & \textbf{Mean (\%)} & \textbf{Std Dev} & \textbf{Min (\%)} & \textbf{Max (\%)} \\
\hline
GB-TWKSVC & 84.67 & 13.83 & 52.44 & 99.34 \\
Twin-KSVC & 81.37 & 13.49 & 54.13 & 97.31 \\
TSVM & 75.10 & 13.46 & 51.17 & 95.33 \\
\hline
\end{tabular}
\caption{Summary statistics of model performance across all datasets}
\label{tab:summary_stats}
\end{table}

GB-TWKSVC achieved the highest mean accuracy (84.67\%), followed by Twin-KSVC (81.37\%) and TSVM (75.10\%). Standard deviations remained consistent across models (13.46-13.83\%).

\subsubsection{Statistical Significance Tests}
Tables \ref{tab:paired_t_test} and \ref{tab:wilcoxon_test} present the results of paired t-tests and Wilcoxon signed-rank tests, respectively.

\begin{table}[h]
\centering
\begin{tabular}{|l|c|c|}
\hline
\textbf{Model Comparison} & \textbf{t-statistic} & \textbf{p-value} \\
\hline
GB-TWKSVC vs Twin-KSVC & 3.347 & 0.0101 \\
GB-TWKSVC vs TSVM & 4.737 & 0.0015 \\
Twin-KSVC vs TSVM & 4.116 & 0.0034 \\
\hline
\end{tabular}
\caption{Results of Paired t-tests for Model Comparisons}
\label{tab:paired_t_test}
\end{table}

\begin{table}[h]
\centering
\begin{tabular}{|l|c|c|}
\hline
\textbf{Model Comparison} & \textbf{W-statistic} & \textbf{p-value} \\
\hline
GB-TWKSVC vs Twin-KSVC & 2.00 & 0.0117 \\
GB-TWKSVC vs TSVM & 0.00 & 0.0039 \\
Twin-KSVC vs TSVM & 0.00 & 0.0039 \\
\hline
\end{tabular}
\caption{Results of Wilcoxon Signed-Rank Tests for Model Comparisons}
\label{tab:wilcoxon_test}
\end{table}

\subsubsection{Discussion of Statistical Results}
The analysis revealed statistically significant differences between all model pairs at the $\alpha = 0.05$ level, with both parametric and non-parametric tests confirming these findings. The most pronounced difference emerged between GB-TWKSVC and TSVM (t = 4.737, p = 0.0015), supported by corresponding Wilcoxon test results. GB-TWKSVC's superior mean accuracy of 84.67\% across datasets, combined with the statistical significance of its performance advantages, establishes it as the optimal choice for practical classification tasks.

\section{Conclusion and Future Work}
This study presents the Granular Ball K-class Twin Support Vector Classifier (GB-TWKSVC), a novel multi-class classification algorithm that combines granular computing principles with TSVM formulations. The core of the GB-TWKSVC model involves partitioning data into granular balls based on hierarchical clustering and K-Means, ensuring that each cluster represents a compact, homogeneous group. These granular balls are used to define decision boundaries through pairwise comparisons between classes, facilitated by solving Quadratic Programming Problems (QPPs) for each class pair. The algorithm's design ensures efficient classification by creating hyperplanes for each class pair and utilizing a voting mechanism for final decision-making.

Experimental evaluation across diverse benchmark datasets demonstrates GB-TWKSVC's effectiveness through superior classification performance compared to state-of-the-art methods. It exhibits enhanced scalability for large-scale problems, with reduced training times, and is robust in handling class imbalance, particularly in datasets with moderate class numbers. The algorithm’s adaptive capacity to local data distributions, supported by its granular computing framework, contributes to its overall effectiveness, establishing GB-TWKSVC as a significant advancement in multi-class classification.

The comprehensive experimental results highlight GB-TWKSVC's practical advantages in accuracy, computational efficiency, and scalability. The algorithm performs consistently across a range of datasets, establishing its applicability in a variety of machine learning domains, particularly those requiring efficient handling of large-scale multi-class classification tasks.

Future research will explore several promising directions. These include the development of online and incremental learning variants for streaming data applications, as well as advanced granular ball construction methods that incorporate feature importance and local density information. The algorithm will also be extended to handle multi-label classification problems. Moreover, there is potential for applying GB-TWKSVC in diverse domains such as computer vision, natural language processing, and bioinformatics. Additional focus will be placed on theoretical analysis, including generalization bounds and convergence properties, and on implementing efficient hyperparameter optimization strategies, potentially using meta-learning or Bayesian approaches. These advancements will further enhance GB-TWKSVC's capabilities and solidify its role in advancing multi-class classification within the broader machine-learning landscape.
\nocite{*}  % This will include all references
\bibliographystyle{unsrt}
\bibliography{references}

\end{document}